\documentclass[journal]{IEEEtran}

\usepackage{amsmath,amssymb,amsfonts}
\usepackage{graphicx}
\usepackage{multirow}
\usepackage{booktabs}
\usepackage{array}
\usepackage{pifont}
\usepackage[utf8]{inputenc}
\usepackage{xcolor}
\usepackage[colorlinks=true,allcolors=blue]{hyperref}
\usepackage{url}

\newcommand{\safeincludegraphics}[2][]{%
  \IfFileExists{#2}{\includegraphics[#1]{#2}}%
  {\fbox{\parbox{0.9\linewidth}{\centering Missing figure file: \texttt{#2}}}}%
}

\begin{document}

\title{Memory-Efficient Active Perception for\\
Embodied Question Answering in Dynamic Human-Populated Scenes}

\author{Xin~Lu, Rui~Li, Xun~Huang, Weixin~Li, Jiayuan~Li,
Chuanqing~Zhuang, Zhengda~Lu, and Jun~Xiao%
\thanks{Affiliations: Xin Lu, Chuanqing Zhuang, Zhengda Lu, and Jun Xiao---
University of Chinese Academy of Sciences; Rui Li and Weixin Li---Beihang
University; Xun Huang---Xiamen University; Jiayuan Li---Beijing Institute of
Technology. Xin Lu, Rui Li, Xun Huang, Weixin Li, and Jiayuan Li are also with
Zhongguancun Academy. Corresponding
authors: Jun Xiao (xiaojun@mails.ucas.ac.cn) and Weixin Li
(weixinli@buaa.edu.cn).}}

\maketitle

\begin{abstract}
Embodied Question Answering (EQA) couples active visual perception,
vision--language reasoning, and sequential decision making, and has traditionally
been evaluated in temporally stable environments where visual evidence can be
accumulated reliably. In realistic multimedia settings such as dynamic
human-populated scenes, however, moving people and transient occlusions introduce
\emph{human-induced perceptual variations}: task-relevant multimodal cues become
transient, viewpoint-dependent, and incompletely captured in individual views.
Conventional store-then-retrieve strategies
generate redundant visual evidence and incur high inference cost, making them
ill-suited to such variations. This exposes two core problems for embodied agents:
resolving viewpoint-dependent ambiguity, and maintaining compact evidence for
efficient inference.
To address these challenges, we present \textbf{DIVRR}
(\textbf{D}ynamic-\textbf{I}nformed \textbf{V}iew \textbf{R}efinement with
\textbf{R}elevance-guided \textbf{R}otation and adaptive memory selection), a
training-free active evidence-acquisition framework for EQA. DIVRR combines
question-conditioned room guidance with relevance-triggered view augmentation to
guide exploration and additional observation. Selective memory admission uses the
same relevance signal as view augmentation to retain high-scoring evidence,
reducing stored entries while keeping the retrieved visual context compact.
To evaluate EQA under human-induced perceptual variations, we introduce
\textbf{DynHuman-EQA}, a dynamic human-populated EQA dataset constructed in
Habitat, comprising dynamic and static subsets under shared scene
layouts. Experiments on DynHuman-EQA and HM-EQA show that DIVRR outperforms existing
baselines in both accuracy and memory efficiency. The gains are most pronounced in
dynamic human-populated scenes.
\end{abstract}

\begin{IEEEkeywords}
  Embodied question answering, active perception, vision--language models,
multi-view reasoning, dynamic human-populated scenes, multimodal memory.
\end{IEEEkeywords}

\section{Introduction}
\label{sec:intro}
\IEEEPARstart{E}{mbodied} Question Answering (EQA) requires an autonomous agent to
explore a 3D environment, actively acquire visual evidence, and generate answers
grounded in the perceived scene~\cite{eqa}. Unlike conventional Visual Question
Answering (VQA), EQA tightly couples active perception, vision--language
understanding, and sequential decision making: the agent must determine
exploration trajectories, inspect informative viewpoints, and answer once
sufficient evidence has been collected~\cite{eqa,yu2019multi-eqa}. As a multimedia
reasoning task, EQA fuses egocentric visual streams, language queries, and spatial
memory over time, and recent foundation models increasingly make such multimodal
fusion feasible while raising the bar for efficient and reliable inference in
realistic embodied settings where both sensing and computation are budgeted~\cite{majumdar2024openeqa}.

A fundamental challenge in EQA lies in balancing perceptual sufficiency and
inference efficiency. Mainstream methods adopt a store-then-retrieve paradigm,
accumulating a large buffer of observations and selecting evidence only at
answering time. This creates a trade-off between coverage and efficiency:
aggressive accumulation improves coverage but causes redundancy and expensive
retrieval, while stricter filtering risks discarding transient yet decisive cues.
Existing pipelines often treat observations as an ever-growing buffer without
explicitly controlling redundancy~\cite{keqa,explore-eqa,seqa}. As exploration
proceeds, repeated similarity checks over highly overlapping views and the
accumulation of low-utility candidates make evidence retrieval progressively more
expensive~\cite{zhai2025memory,msgnav}. Although structured representations such as
3D semantic scene graphs can reduce duplication~\cite{saxena2024grapheqa}, they
incur substantial perception and maintenance overhead: per-frame detection or
segmentation, RGB-D processing, 3D fusion, and incremental graph updates must run
online, which limits scalability for lightweight embodied agents operating in
large and cluttered indoor scenes.

\begin{figure}[!t]
  \centering
  \safeincludegraphics[width=\linewidth]{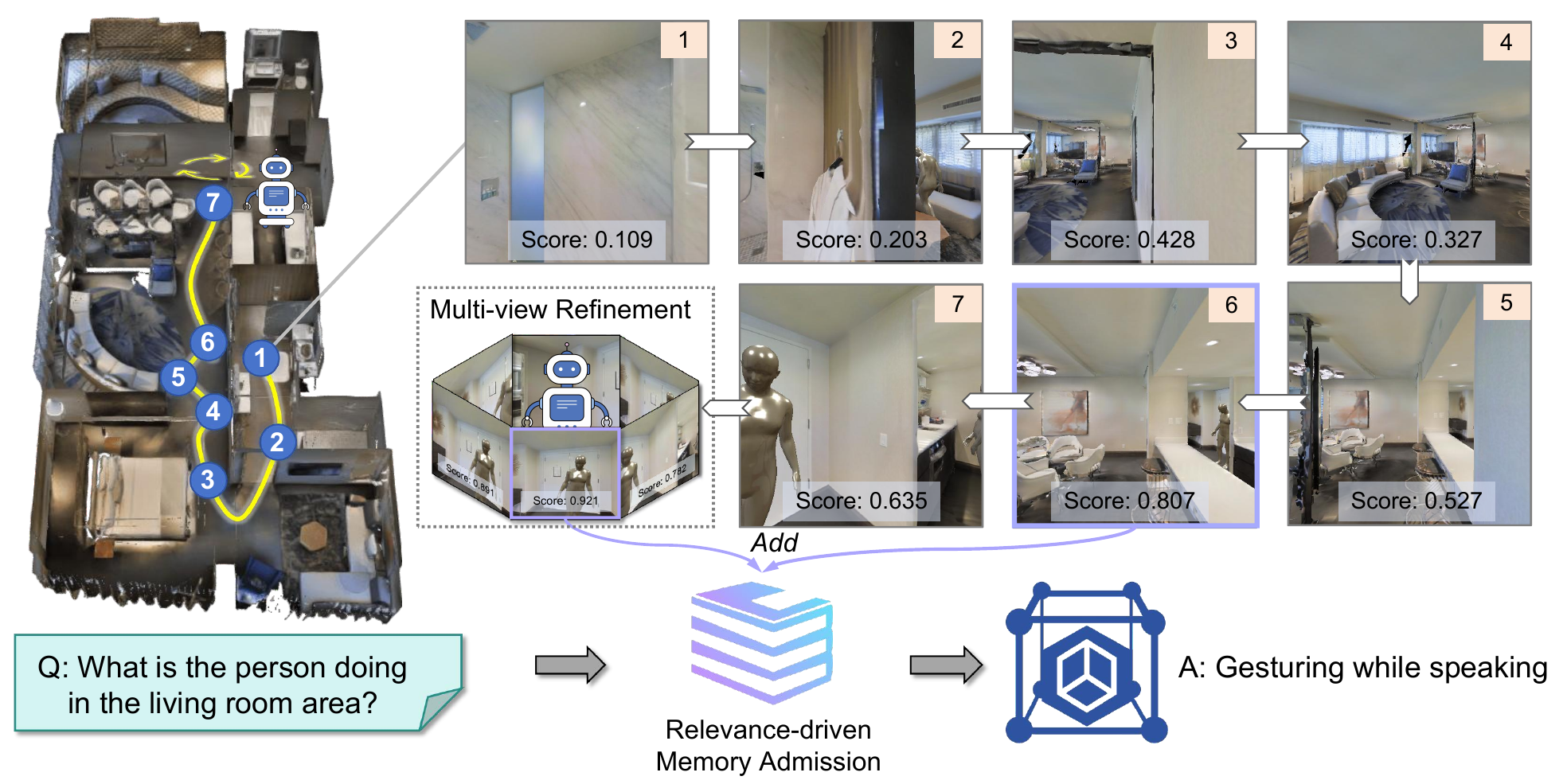}
  \caption{Evidence acquisition in \textbf{DIVRR}. Values denote relevance scores
  $s_t$; purple boxes mark observations admitted to memory. At Step~7 the person is
  incompletely observed in the current view; its relevance score falls in the
  refinement band, triggering multi-view
  augmentation to inspect additional directions before committing evidence.}
  \label{fig:sample}
\end{figure}

This challenge becomes more pronounced in dynamic human-populated scenes, where
moving people introduce transient occlusions, visibility changes, and
view-sensitive cues~\cite{dong2025ha}. We refer to this condition as
\emph{human-induced perceptual variations}: the agent faces locally ambiguous and
time-varying observations caused by human motion. Occlusions hide portions of the
scene, while limited angular coverage and partial target framing can also leave
relevant content unobserved. As a concrete example
(Fig.~\ref{fig:sample}), answering ``\emph{what is the person doing in the living
room?}'' can fail when the current view captures only part of the person's activity:
the available evidence supports several activity labels. Critical
evidence is often visible only briefly and from specific perspectives, so simple
coverage expansion alone may miss them; the agent also needs to inspect
question-relevant views during exploration.

To study this setting, we introduce \textbf{DynHuman-EQA}, a Habitat-based dynamic
human-populated EQA dataset with dynamic and static subsets drawn from shared
scene layouts. Unlike prior EQA datasets that mainly
assume stable observations, DynHuman-EQA is explicitly designed to evaluate
multi-view evidence acquisition under viewpoint-dependent visibility.

Built on this setting, we propose \textbf{DIVRR}, a training-free active
evidence-acquisition framework for EQA with two complementary components.
For exploration, room detection (RD) provides question-conditioned room guidance,
while view augmentation allocates additional observations to intermediate-relevance
views (Fig.~\ref{fig:sample}). For memory, selective admission retains high-scoring
evidence rather than every acquired view. The same relevance score triggers view
augmentation, ranks its candidates, and tests the selected view for admission,
linking evidence acquisition to storage without retaining unselected auxiliary views.
Comparisons on DynHuman-EQA show accuracy gains of \textbf{10.10}
and \textbf{4.00} points on the dynamic and static subsets over their strongest
baselines, while reducing admitted memory entries by \textbf{94\%}
relative to MemoryEQA on the dynamic subset.
The main contributions of this work are summarized as follows:
\begin{itemize}
  \item We introduce \textbf{DynHuman-EQA}, a dynamic human-populated EQA dataset
  with dynamic and static subsets drawn from shared scene layouts, supporting
  evaluation of embodied multimodal reasoning with incomplete visual evidence
  across viewing directions. We describe its generation and quality-control
  protocol in the supplementary material.
  \item We develop question-guided exploration in \textbf{DIVRR}, combining
  room detection to guide exploration toward task-relevant regions with
  relevance-triggered view augmentation to inspect additional directions at
  intermediate-relevance observations.
  \item We introduce selective memory admission within DIVRR to reduce stored
  observations. Using the same relevance signal as view augmentation, it admits
  at most one high-scoring, valid observation per waypoint and excludes
  unselected auxiliary views from memory.
  \item We conduct extensive experiments on DynHuman-EQA and HM-EQA, including
  baseline comparisons, active-perception controls, and module ablations,
  showing that DIVRR improves answer accuracy and memory efficiency over existing
  baselines, with particularly clear gains in
  dynamic human-populated scenes.
\end{itemize}

\section{Related Work}
\label{sec:related}

\subsection{EQA Datasets}
Unlike static 2D VQA, EQA requires agents to reason within 3D environments while
actively acquiring visual evidence, making dataset construction more challenging.
Early EQA datasets~\cite{videonavqa,gordon2018iqa,wijmans2019embodied} often rely
on template-based generation, which enables scalable collection but produces rigid
question formats and limited scene complexity. Recent works leverage foundation
models or manual design to improve linguistic diversity and task realism,
including S-EQA~\cite{seqa}, Explore-EQA~\cite{explore-eqa},
NoisyEQA~\cite{wu2024noisyeqa}, OpenEQA~\cite{majumdar2024openeqa},
CityEQA~\cite{zhao2025cityeqa}, and MT-HM3D~\cite{zhai2025memory}.

Despite these advances, existing EQA datasets are still largely built around
temporally stable observations, where evidence can be reliably accumulated once
discovered. Human-aware navigation research~\cite{dong2025ha} highlights the
importance of moving people and occlusion-aware perception, yet current EQA
datasets rarely evaluate how human-induced occlusions and viewpoint-dependent
visibility affect evidence acquisition and downstream reasoning in dynamic
human-populated scenes. The most related effort is \emph{Enter the Mind
Palace}~\cite{mindpalace}, which studies \emph{long-term} active EQA under
episodic environmental variation across visits. Our setting is complementary:
DynHuman-EQA targets \emph{within-episode} human motion, transient occlusion, and
viewpoint-dependent evidence, and provides static/dynamic subsets under
shared layouts, with question content and observation sampling tailored to
each subset.

\subsection{Multimodal Question Answering and Evidence Selection}
\label{sec:related_mmqa}
EQA inherits much of its machinery from multimedia question answering, where the
central difficulty is likewise deciding \emph{which} visual evidence supports a
language query. In image-based VQA, relational reasoning over object-level
representations strengthens the features a question attends
to~\cite{yu2020reasoning}, and causal formulations discourage models from exploiting
language priors over the image~\cite{pan2024unbiased}. Video QA makes selection
explicit, since the answer usually depends on a small part of a long stream: graph
reasoning units disentangle appearance from motion~\cite{wang2022dualvgr},
lightweight architectures pursue the same goal under tighter
budgets~\cite{jiang2023livlr}, others first localize the temporal window the
question implies~\cite{qian2024locate}, and multimodal large language models have
been adapted by identifying question-critical, temporally coherent moments worth
conditioning on~\cite{wang2025adapting}.

These works share our premise that selective, question-conditioned evidence beats
indiscriminate accumulation, but differ in one decisive respect: their input stream
is given in advance, so selection is \emph{post hoc} filtering over material the
model already holds. An embodied agent must instead \emph{act} to bring decisive
evidence into view, making acquisition an integral part of evidence selection.

\subsection{Models for Embodied Agents in EQA}
Foundation models have recently advanced embodied tasks such as vision--language
navigation~\cite{long2024instructnav,long2024discuss,zheng2024generalist,zhang2025flexvln,wu2024latent,li2026dcsen},
robotic manipulation~\cite{kapelyukh2024dream2real,wu2023tidybot,xu2024rtgrasp}, and
EQA~\cite{eqa,gordon2018iqa,yu2019multi-eqa}. Two concerns from the navigation
literature carry over directly to our setting. The first is how an agent should
combine what it currently sees with what it has stored, which continuous-navigation
systems address by making memory and observation cooperate rather than treating the
buffer as a passive log~\cite{yu2025mossvln}. The second is what an agent should do
when its own predictions are unreliable, which zero-shot navigation methods handle by
estimating uncertainty and acting on it~\cite{shi2026reliablenav}. Existing EQA methods use these models
for high-level task decomposition, target-object detection~\cite{sakamoto2024map},
hierarchical planning~\cite{zhao2025cityeqa}, exploration-direction
scoring~\cite{cheng2024efficienteqa,explore-eqa}, or dense semantic-map
construction~\cite{saxena2024grapheqa}. Large language models have also been used
to supply contextual priors that guide an embodied questioner toward plausible
target locations~\cite{wang2026contextaware}. Closest in spirit to our approach is
target-focused perception, which concentrates the agent's perceptual effort on
question-relevant targets rather than on uniform
exploration~\cite{wang2026targetfocused}. Memory-centric EQA organizes and retrieves
past observations to support exploration and answering~\cite{zhai2025memory}.
DIVRR focuses on the decision before memory writing: a shared relevance score
links the need for another view to the criterion for retaining it. This conditional
acquisition-and-admission rule complements question-conditioned room guidance for
exploration, with at most one selected memory entry per waypoint.

Memory for embodied multimodal reasoning spans recurrent states and metric maps
~\cite{graves2012lstm,szot2021habitat2,wijmans2019ddppo,anderson2019chasing,blukis2018mapping,cartillier2021semantic,chaplot2020object},
topological and semantic graphs~\cite{savinov2018semi,wu2019bayesian,cheng2018geometry,prabhudesai2019embodied,tung2019learning,saxena2024grapheqa},
and retrieval-augmented stores~\cite{anwar2024remembr,xie2024embodiedrag,cheng2024efficienteqa}.
These representations motivate selective retention of task-relevant evidence from incomplete observations.

\section{The DynHuman-EQA Dataset}
\label{sec:dataset}

We introduce \textbf{DynHuman-EQA}, a Habitat-based EQA dataset for evaluating
embodied multimodal reasoning under human-induced perceptual dynamics. Unlike prior
EQA datasets, which predominantly assume temporally stable observations, it
introduces moving people and motion-induced occlusions that make task-relevant cues
transient and viewpoint-dependent. Individual views may provide incomplete evidence:
occlusions hide parts of the scene, and the current field of view may omit relevant
directions or capture only part of a target. It comprises dynamic and static subsets drawn
from shared scene layouts, with question content and observation sampling
tailored to each subset. Here,
``\emph{temporally stable}'' means the observation does not change over time at a
fixed pose.
The two subsets evaluate stable and human-populated observations with their
respective question distributions.

\begin{figure}[!t]
  \centering
  \safeincludegraphics[width=0.92\linewidth]{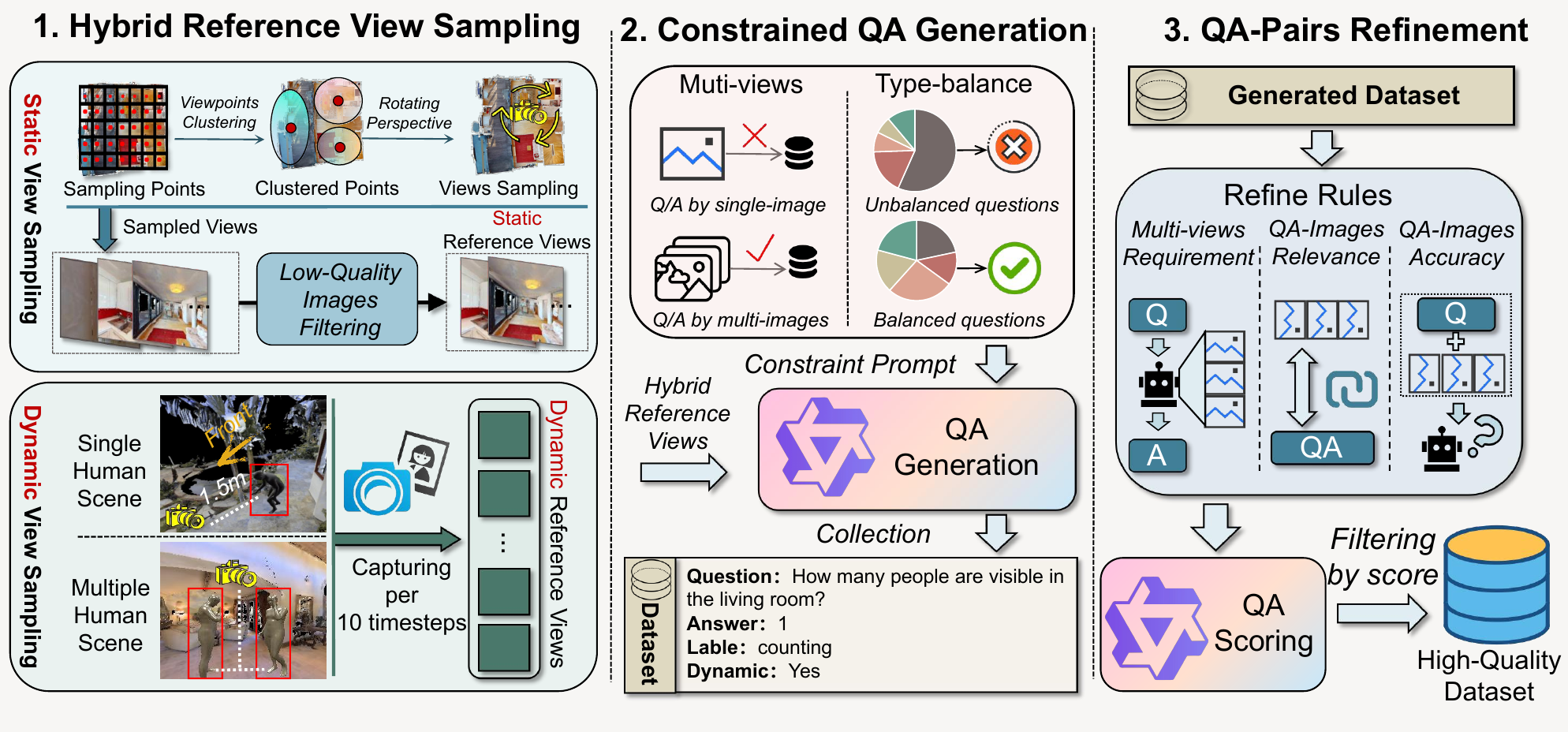}
  \caption{Construction pipeline of DynHuman-EQA. We build static and
  dynamic subsets under shared scene layouts, generate QA pairs from multi-view observations with
  Qwen3-VL-Plus, and manually review candidate QA pairs to obtain
  the final dataset.}
  \label{fig:dataset_pipeline}
\end{figure}

\subsection{Construction pipeline}
As shown in Fig.~\ref{fig:dataset_pipeline}, DynHuman-EQA follows a unified
construction pipeline that first samples observation locations from navigable
regions and then instantiates either stable or human-populated observations
according to the subset definition. For the static subset, we cluster spatial
points and render multiple rotated room views under temporally stable conditions.
For the dynamic subset, we capture observations along human motion sequences: we
place observation points 1.5\,m in front of a person's trajectory, or along the
perpendicular bisector between interacting individuals when their distance is
below 2.0\,m, and capture frames every 10 timesteps from 120-frame motion
sequences. This introduces transient occlusions and viewpoint-dependent
visibility, yielding observations with varying coverage of task-relevant content.
The same scene layout can thus be evaluated under both stable and
dynamic perceptual conditions over identical underlying geometry.

\subsection{Question generation, filtering, and human verification}
\label{sec:dataset_qa}
Question--answer pairs are generated from multi-view observations rather than
single images, encouraging agents to synthesize evidence across distinct
viewpoints. We use \emph{Qwen3-VL-Plus} \emph{only} for offline question
generation; inference-time relevance scoring and answer generation use the local
Qwen2.5-VL-7B backbone by default, with separate backbone-sensitivity experiments
(Sec.~\ref{sec:experiments}).

\paragraph*{Generation constraints}
Question generation is governed by explicit format constraints, which differ
between the two subsets. For the \textbf{dynamic subset}, questions are constructed
from sequential observations of the same person and are required to: (1) avoid
referring to the images directly; (2) exploit temporal observations rather than a
single frame; (3) exclude unobservable internal states such as emotions;
(4) balance labels across the state, object, existence, and location categories;
(5) exclude ``why'' and ``how'' questions, which invite speculation rather than
perception; (6) avoid ambiguous positional descriptors; (7) provide four answer
options with exactly one definitive answer; and (8) for object questions, refer to
the object the person attends to rather than the object merely held in hand. For
the \textbf{static subset}, questions are generated from multi-view room
observations and are required to: (1) avoid referring to the images directly;
(2) require multi-view evidence wherever possible; (3) admit only visible static
states, such as a light being on or off, while excluding dynamic or invisible
properties; (4) balance labels across the object, existence, attribute, location,
state, and counting categories; (5) exclude ``why'' and ``how'' questions;
(6) avoid ambiguous positional descriptors; and (7) provide four answer options
with a definitive answer. Both formats encourage cross-view reasoning. To counteract the natural
drift toward easily satisfiable question types such as counting, we track
per-category generation counts and embed the running statistics directly into the
prompt that was used for generation.

\paragraph*{Filtering and human verification}
We \emph{manually review} candidate items for question quality, answerability, and
relevance to the source observations, favoring questions that encourage cross-view
reasoning. The review removes low-quality items, including trivial partial-view
questions and those based on degraded renders. The complete generated pool contains
1{,}225 candidate QA pairs, comprising 612 human-populated and 613 static instances;
manual quality filtering retains the 1{,}100 reported here, a retention rate of
89.8\%. We evaluate the retained question--answer pairs using exact match against
their \emph{human-verified answer keys}. The supplementary material summarizes the generation constraints
for both subsets and provides representative filtering illustrations.

\subsection{Statistics and comparison}
DynHuman-EQA contains 1{,}100 question--answer pairs drawn from 62
Matterport3D~\cite{chang2017matterport3d} scenes, evenly split between 550 dynamic and
550 static questions, spanning seven categories: attribute, counting, existence,
interaction, location, object, and state. Per-category statistics are reported in the
supplementary material.

Table~\ref{tab:dataset_comparison} positions DynHuman-EQA against widely used EQA
benchmarks along four axes that matter here: whether scenes are reconstructed from
real environments, whether observations contain moving people, whether ground-truth
evidence poses are recorded, and how questions are created. Early
benchmarks~\cite{eqa,wijmans2019embodied,yu2019multi-eqa,gordon2018iqa} rely on
rule-based templates, which scale well but constrain linguistic variety, while more
recent ones~\cite{fine-eqa,explore-eqa,seqa,zhao2025cityeqa,zhai2025memory,majumdar2024openeqa}
use foundation models or manual curation to broaden diversity. None of them populate
the scene with moving people. To our knowledge, DynHuman-EQA is therefore the first
EQA dataset combining real 3D scenes, explicitly dynamic human-populated
observations, and recorded ground-truth evidence poses.

\begin{table}[!t]
  \centering
  \caption{Comparison of DynHuman-EQA with existing EQA datasets. \textbf{Real}
  indicates scenes reconstructed from real environments, \textbf{Dynamic}
  indicates moving people in the observations, and \textbf{GT obs.} indicates
  recorded ground-truth evidence poses.}
  \label{tab:dataset_comparison}
  \setlength{\tabcolsep}{3pt}
  \resizebox{\linewidth}{!}{%
  \begin{tabular}{l|cccccc}
    \toprule
    \textbf{Dataset} & \textbf{Simulator} & \textbf{Source} & \textbf{Real} & \textbf{Dynamic} & \textbf{GT obs.} & \textbf{Creation} \\
    \midrule
    EQA-v1~\cite{eqa}                   & House3D      & SUNCG        & No  & No  & No  & Rule \\
    MP3D-EQA~\cite{wijmans2019embodied} & MINOS        & MP3D         & Yes & No  & Yes & Rule \\
    MT-EQA~\cite{yu2019multi-eqa}       & House3D      & SUNCG        & No  & No  & Yes & Rule \\
    IQA~\cite{gordon2018iqa}            & AI2THOR      & --           & No  & No  & No  & Rule \\
    EXPRESS-Bench~\cite{fine-eqa}       & Habitat      & HM3D         & Yes & No  & Yes & VLMs \\
    HM-EQA~\cite{explore-eqa}           & Habitat      & HM3D         & Yes & No  & No  & VLMs \\
    S-EQA~\cite{seqa}                   & VirtualHome  & --           & No  & No  & No  & LLMs \\
    CityEQA~\cite{zhao2025cityeqa}      & EmbodiedCity & --           & No  & No  & Yes & Manual \\
    MT-HM3D~\cite{zhai2025memory}       & Habitat      & HM3D         & Yes & No  & No  & VLMs \\
    Open-EQA~\cite{majumdar2024openeqa} & Habitat      & ScanNet/HM3D & Yes & No  & No  & Manual \\
    \midrule
    \textbf{DynHuman-EQA}               & Habitat      & MP3D         & Yes & Yes & Yes & VLMs \\
    \bottomrule
  \end{tabular}%
  }
\end{table}

The two subsets let one pipeline be evaluated under both stable and human-induced
dynamic observations, and recording evidence poses enables the geometric reliability
analysis of Sec.~\ref{sec:reliability}.

DynHuman-EQA evaluates active EQA: the Habitat agent explores the environment,
updates its memory, and answers from self-acquired observations.

\section{The DIVRR Framework}
\label{sec:method}

\begin{figure*}[!t]
  \centering
  \safeincludegraphics[trim=0 69.24 0 31.26,clip,width=0.82\textwidth]{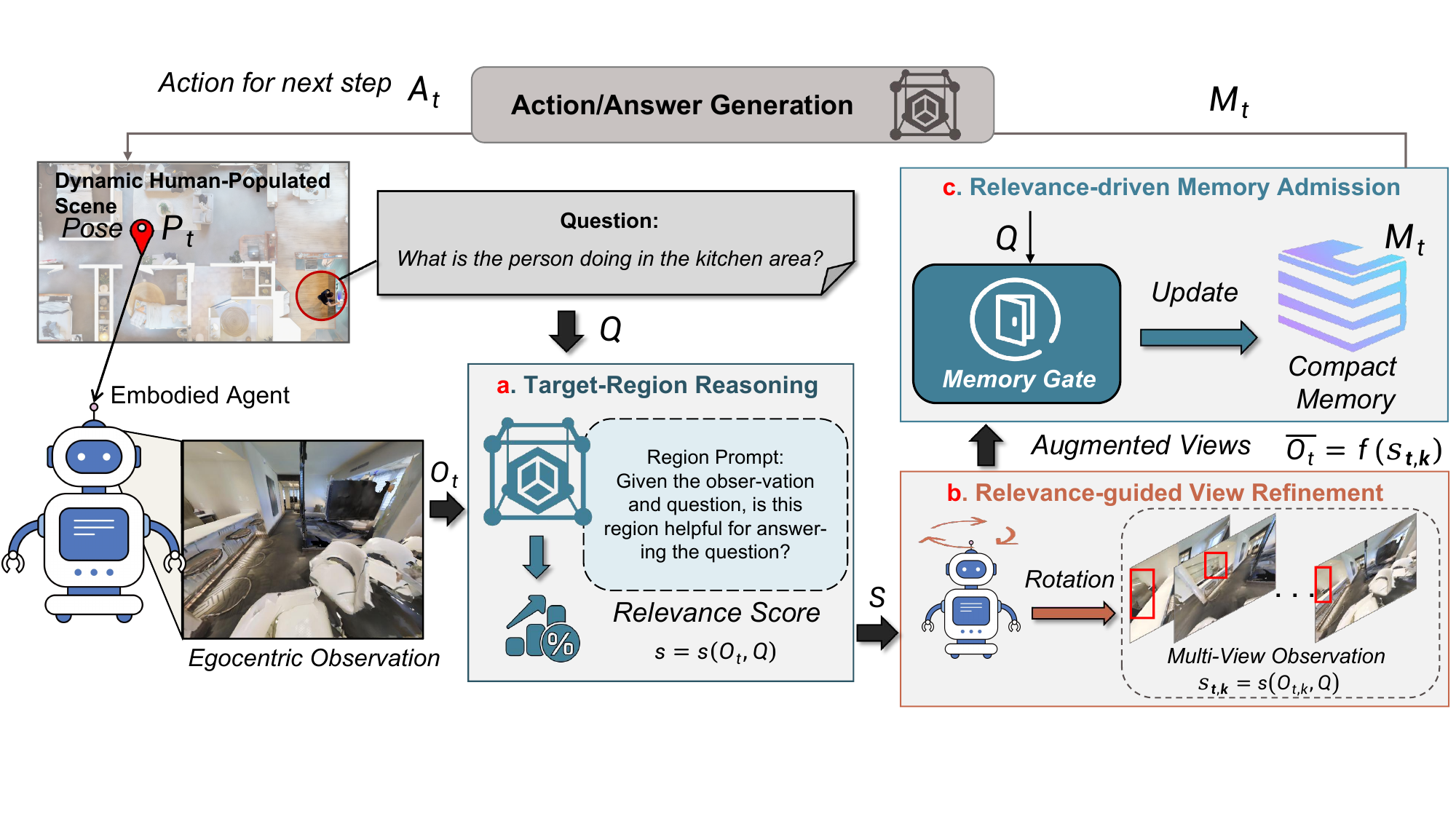}
  \caption{DIVRR for a kitchen-area activity question, coupling
  \textbf{(a)} target-region reasoning, \textbf{(b)} relevance-guided multi-view
  refinement, and \textbf{(c)} relevance-driven memory admission under a single
  question-conditioned relevance signal.}
  \label{fig:overview}
\end{figure*}

\textbf{DIVRR} combines question-guided exploration with selective memory admission.
RD guides the exploration mode, while view augmentation provides additional
observations at selected waypoints; admission controls which observations are stored.
Figure~\ref{fig:overview} illustrates the question
``\emph{What is the person doing in the kitchen area?}'' At step $t$ the
agent is at pose $P_t$ and receives an egocentric observation $O_t$. Because a single
view may omit relevant directions or capture a target incompletely,
an unfiltered accumulation strategy admits redundant or partially informative
evidence. Given the question $Q$, DIVRR uses a vision--language model (VLM) for
target-region reasoning, obtaining a question-conditioned relevance score $s_t$. When $s_t$ falls in the intermediate-score refinement band, DIVRR
performs relevance-guided view refinement via multi-view augmentation, acquiring a
small set of auxiliary views $\overline{O}_t$ and selecting the highest-scoring view
$\widetilde{O}_t$. Finally, relevance-driven memory admission decides whether to
write $\widetilde{O}_t$ into memory $M_t$ based on its relevance score and image
validity. The updated memory is fed back to the explorer
for next-step action $A_t$ and to the VLM for final answer generation.

\subsection{Exploration pipeline}
\label{sec:roles}
DIVRR builds on a frontier-based
explorer (FBE), where a local scoring prompt (LSV) ranks candidate frontier
directions conditioned on the question and a global scoring prompt (GSV) decides
whether any direction in the current view is worth exploring. DIVRR adds a
room-detection gate (Sec.~\ref{sec:rd}) that selects between goal-oriented and
frontier-based exploration, together with relevance-triggered view augmentation.
Its memory-admission stage retains selected high-scoring observations. Action selection is
governed by a fixed rule $A_t=\pi(O_t,Q,M_t)$ over these components.

\subsection{Room detection}
\label{sec:rd}
Room detection (RD) supplies a discrete semantic gate that selects the exploration
mode. At the start of an episode, the VLM maps the question to an ordered subset
$\mathcal{R}_Q$ of potentially relevant room types drawn from 20 predefined
categories (e.g.\ bathroom, kitchen, living room, hallway, and a catch-all
\emph{unknown}). At each waypoint the same VLM classifies the current RGB
observation over these categories using option-token probabilities. Because
single-frame predictions fluctuate in doorways and transitional spaces, we keep the
three most recent predictions and combine them by confidence-weighted voting,
weighting recent frames more. Writing $\widehat{r}_t$ for the smoothed room label, the
exploration gate is $d_t=\mathbb{I}[\widehat{r}_t\in\mathcal{R}_Q]$.
When $d_t=1$ the agent enables goal-oriented exploration toward the predicted
task-relevant region; otherwise it falls back to frontier-based exploration. RD thus
contributes a question-conditioned room prior together with a temporally smoothed
estimate of the current room to guide exploration toward task-relevant regions.

\subsection{Target-region reasoning}
\label{sec:tr_reasoning}
DIVRR treats question answering as evidence acquisition under a limited sensing
budget, where the key challenge is to distinguish question-relevant cues from
redundant or uninformative observations. Given $O_t$ and $Q$, we query the VLM
with a target-region reasoning prompt asking whether the current observation
contains information useful for answering $Q$. The VLM produces a scalar relevance
score $s_t\in[0,1]$ that serves as a unified signal for both view refinement and
memory admission.

\paragraph*{Zero-shot relevance score}
Following prior zero-shot EQA protocols, we extract the next-token logits from the
VLM. Let $\mathbf{z}_t\in\mathbb{R}^{|\mathcal{V}|}$ be the logits under the
target-region reasoning prompt conditioned on $(O_t,Q)$. Let $\mathcal{T}^{+}$ be
affirmative tokens (e.g., ``Yes''), and let $\mathcal{V}_{\mathrm{cand}}$ be the
candidate set of polar response tokens. With token-level evidence
$E_t(w)=\mathbf{z}_t[\mathrm{id}(w)]$, we compute
\begin{equation}
  s_t = s(O_t,Q) =
  \frac{\sum_{w\in\mathcal{T}^{+}}\exp\!\left(E_t(w)/\tau\right)}
       {\sum_{w'\in\mathcal{V}_{\mathrm{cand}}}\exp\!\left(E_t(w')/\tau\right)},
  \label{eq:relevance_score}
\end{equation}
where $\tau$ is a temperature coefficient.

\subsection{Relevance-guided view refinement}
\label{sec:view_refine}
An observation may contain question-relevant cues without providing a sufficiently
informative view of the target. DIVRR performs view refinement through bounded multi-view augmentation
before memory writing.

\paragraph*{Triggering rule}
View refinement is triggered when the current observation's relevance score
falls in the refinement band between the refinement
threshold $\tau_{\mathrm{rot}}$ and the memory threshold $\tau_{\mathrm{mem}}$:
\begin{equation}
  \tau_{\mathrm{rot}}\le s_t < \tau_{\mathrm{mem}},
  \label{eq:mv_trigger}
\end{equation}
Observations below $\tau_{\mathrm{rot}}$ bypass refinement and admission, and those
at or above $\tau_{\mathrm{mem}}$ become direct admission candidates, allocating
additional observations to the intermediate-score range.

When refinement is not triggered, we set $\widetilde{O}_t=O_t$ for retrieval.
The current view becomes an admission candidate only when
$s_t\ge\tau_{\mathrm{mem}}$; views below $\tau_{\mathrm{rot}}$ are not admitted.
Every admission candidate passes through the same validity check before memory
writing.

\paragraph*{Multi-view augmentation and candidate selection}
Once triggered, the agent performs multi-view augmentation and collects a bounded
set of auxiliary observations
$\overline{O}_t=\{O_{t,k}\}_{k=1}^{N}$, where $N$ is a fixed sensing budget, i.e.\ the
maximum number of auxiliary views per waypoint. Each $O_{t,k}$ samples an additional
viewing direction. We then re-evaluate relevance
for each view and select the highest-scoring candidate
\begin{equation}
  k^{\star}=\arg\max_{k\in\{1,\dots,N\}} s(O_{t,k},Q),
  \qquad \widetilde{O}_t=O_{t,k^\star}.
  \label{eq:select_verified}
\end{equation}
Selecting $\widetilde{O}_t$ before admission keeps unselected auxiliary views out of memory.
For observations in the refinement band, view augmentation inspects additional
directions and can improve target framing, supplying candidates for evidence
selection under the same relevance criterion. Section~\ref{sec:active_controls} compares
the resulting view-selection policy with random views and waiting under a common
extra-observation cap.

\begin{figure}[!t]
  \centering
  \safeincludegraphics[width=0.88\linewidth]{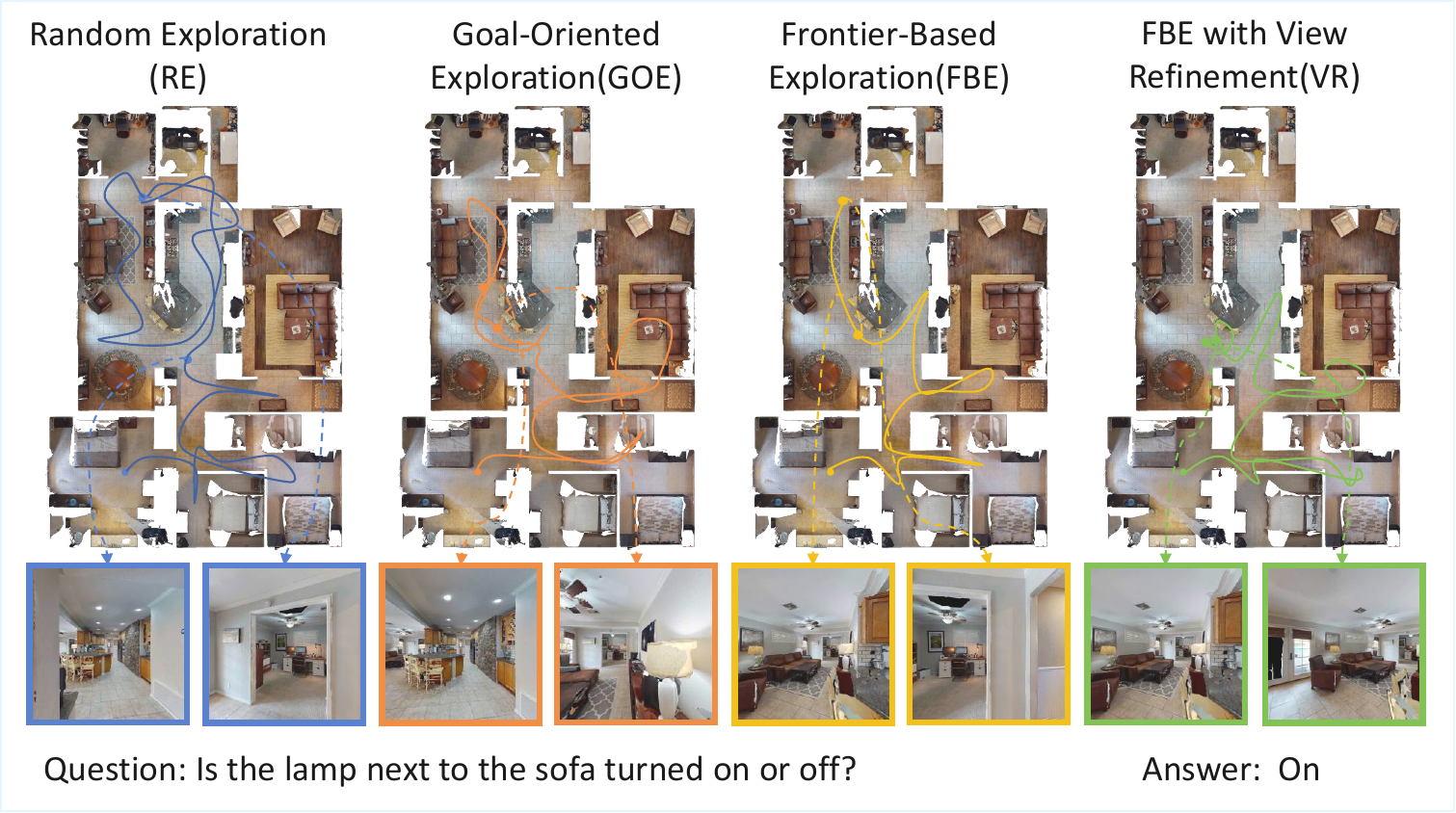}
  \caption{Comparison of exploration trajectories. DIVRR combines exploration
  with relevance-guided inspection of additional views at selected waypoints.}
  \label{fig:explore_strategy}
\end{figure}

\subsection{Selective memory management through relevance-driven admission}
\label{sec:mem_admit}
DIVRR maintains a memory $M_t$ that stores only compact and informative evidence,
which is important in dynamic scenes where redundant and uninformative views
can quickly increase retrieval cost. Memory is updated at most once per waypoint
using the selected view $\widetilde{O}_t$.

\paragraph*{Admission gate}
An observation $\widetilde{O}_t$ is admitted only when it is both highly relevant and
valid:
\begin{equation}
  g_t = \mathbb{I}\!\left[\, s(\widetilde{O}_t,Q)\ge\tau_{\mathrm{mem}}
  \ \wedge\ \mathrm{Valid}(\widetilde{O}_t)\, \right],
  \label{eq:mem_gate}
\end{equation}
where $\mathrm{Valid}(\cdot)$ is a lightweight image-quality test: an observation is
rejected when its fraction of black pixels exceeds a fixed ratio, which discards
degenerate renders where the camera is flush against geometry or sees outside the
reconstructed mesh. Such frames carry no usable evidence yet would occupy a memory
slot. When $g_t=1$, we write a compact memory entry
derived from $\widetilde{O}_t$; otherwise memory is unchanged.

\paragraph*{Entry representation and update}
An admitted entry is $m_t=(\mathbf{m}_t,c_t,\widetilde{P}_t,\widetilde{O}_t)$: the selected view
itself, a textual description $c_t$, its actual capture pose $\widetilde{P}_t$
(including camera orientation), and a retrieval vector
$\mathbf{m}_t$ defined below, giving the update
\begin{equation}
  M_t=
  \begin{cases}
    M_{t-1}\cup\{m_t\}, & \text{if } g_t=1,\\
    M_{t-1}, & \text{otherwise.}
  \end{cases}
  \label{eq:mem_update}
\end{equation}
Each entry retains its image and description alongside its retrieval embedding.
Admission control admits at most one entry per waypoint.
Accordingly, \textbf{Mem} is the average number of admitted
entries per question.

\paragraph*{Memory retrieval}
DIVRR retrieves up to
$K_{\mathrm{ret}}$ entries per step. For an admitted entry we encode the view and its description with
CLIP ViT-L/14 and concatenate the two 768-D features into a retrieval vector
$\mathbf{m}_i=[f_I(\widetilde{O}_i);f_T(c_i)]\in\mathbb{R}^{1536}$, where
$[\cdot\,;\cdot]$ denotes concatenation. The vectors are held in a FAISS
\textsc{IndexFlatL2} index. At step $t$ the query
$\mathbf{q}_t=[f_I(\widetilde{O}_t);f_T(Q)]$ is formed the same way from the current
selected view and the question prompt, and entries are ranked by ascending squared
Euclidean distance to it, with $K_{\mathrm{ret}}{=}5$ throughout. The retrieved
images and descriptions supply the evidence used for answering.

\subsection{Action and final answer generation}
\label{sec:act_answer}
DIVRR uses the evolving memory for exploration and final answering. The explorer
selects the next action as $A_t=\pi(O_t,Q,M_t)$. After exploration, the VLM receives
the question, the answer options, the current selected view, and the
$K_{\mathrm{ret}}$ retrieved entries with their images and textual descriptions,
then selects an option letter. With $K_{\mathrm{ret}}{=}5$, the prediction call
therefore conditions on at most six images. Retrieved entries provide evidence
from earlier waypoints for final answering.

\begin{table*}[!t]
  \vspace{-4pt}
  \centering
  \caption{Comparison on DynHuman-EQA: mean$\pm$sample standard deviation over
  three seeds. Acc: accuracy (\%, $\uparrow$); Mem: admitted entries per question
  ($\downarrow$); Time: latency (s, $\downarrow$). Bold and underlined entries
  indicate the best and second-best means; ``--'' denotes no explicit memory store.}
  \label{table:main}
  \setlength{\tabcolsep}{7pt}
  \scriptsize
  \begin{tabular}{lccc|ccc}
    \toprule
    & \multicolumn{3}{c|}{\textbf{dynamic}} & \multicolumn{3}{c}{\textbf{static}} \\
    \textbf{Method} & \textbf{Acc} & \textbf{Mem} & \textbf{Time} & \textbf{Acc} & \textbf{Mem} & \textbf{Time} \\
    \midrule
    Explore-EQA & $42.03\pm0.35$ & -- & $\mathbf{5.510\pm0.069}$ & $50.50\pm0.30$ & -- & $\mathbf{4.994\pm0.044}$ \\
    Fine-EQA & $\underline{45.00\pm0.40}$ & -- & $5.797\pm0.064$ & $51.30\pm0.40$ & -- & $6.151\pm0.067$ \\
    Graph-EQA & $44.23\pm0.45$ & $\underline{17.30\pm0.40}$ & $14.256\pm0.256$ & $\underline{54.23\pm0.55}$ & $\underline{25.80\pm0.60}$ & $12.912\pm0.259$ \\
    MemoryEQA & $29.80\pm0.80$ & $73.53\pm1.30$ & $30.156\pm0.557$ & $52.47\pm0.55$ & $38.47\pm0.65$ & $24.831\pm0.382$ \\
    DIVRR & $\mathbf{55.10\pm0.40}$ & $\mathbf{4.50\pm0.20}$ & $\underline{5.705\pm0.061}$ & $\mathbf{58.23\pm0.45}$ & $\mathbf{14.40\pm0.40}$ & $\underline{5.793\pm0.062}$ \\
    \bottomrule
  \end{tabular}
  \vspace{-4pt}
\end{table*}

\section{Experiments}
\label{sec:experiments}

\subsection{Implementation details}
\label{sec:impl}
For the default configuration and main baseline comparisons, all VLM-based scoring
and answer generation use the same Qwen2.5-VL-7B backbone; methods differ only in
exploration, reasoning, and memory mechanisms. The separate backbone-sensitivity
experiment in Table~\ref{table:vlm_ablation} replaces this default backbone within
DIVRR. Admitted observations and their textual descriptions are encoded with
CLIP ViT-L/14~\cite{radford2021learning} into 768-D features that are concatenated
into the 1536-D retrieval vector of Sec.~\ref{sec:mem_admit}, following
Memory-EQA~\cite{zhai2025memory}. We follow the
Explore-EQA~\cite{explore-eqa} setup in Habitat-Sim~\cite{savva2019habitat} and set
$\tau{=}1$, $\tau_{\mathrm{mem}}{=}0.8$, $\tau_{\mathrm{rot}}{=}0.6$,
$N{=}3$ auxiliary viewpoints per waypoint, and
$K_{\mathrm{ret}}{=}5$ retrieved memory entries.
The camera's horizontal FOV is $120^{\circ}$, and the verification FOV for
auxiliary views and pose-grounded analysis is $110^{\circ}$.
All inference runs on a single NVIDIA A800 (40\,GB) GPU. \textbf{Mem} denotes the
average number of admitted memory entries per question; methods without an explicit
memory store are marked ``--''. \textbf{Step} in Table~\ref{table:hmeqa} denotes the
average number of extra sensing steps taken per waypoint. A complete hyper-parameter
listing appears in the supplementary material. Tables~\ref{table:main} and
\ref{tab:active-controls} report means and sample standard deviations across
seeds 42, 39, and 45 ($n=3$); the remaining tables and supplementary category
analysis use seed 42.

\subsection{Baseline comparison setup}
\label{sec:fairness}
All methods in the baseline comparisons share the Qwen2.5-VL-7B backbone, aligned
prompts, and equal exploration horizons and VLM-call budgets; each baseline's own settings are kept
fixed. We evaluate Explore-EQA's adaptation from GPT-4V to the shared backbone on HM-EQA:
Qwen2.5-VL-7B raises its accuracy from 50.3\% (GPT-4V) to 54.6\%.
With the shared backbone, DIVRR reaches 63.8\%, a 9.2-point gain over Explore-EQA.

\subsection{Comparison with existing methods}
\label{sec:sota_compare}
We compare DIVRR with representative EQA methods: semantic frontier-based
exploration~\cite{explore-eqa}, goal-oriented exploration~\cite{fine-eqa},
structured 3D scene-graph reasoning~\cite{saxena2024grapheqa}, and
memory-centric retrieval~\cite{zhai2025memory}.

\paragraph*{Results on DynHuman-EQA}
Table~\ref{table:main} shows that DIVRR achieves the highest mean accuracy on both
subsets: $55.10\pm0.40$\% dynamic and $58.23\pm0.45$\% static, exceeding the
strongest baselines, Fine-EQA and Graph-EQA, by \textbf{10.10} and \textbf{4.00}
percentage points. On the dynamic subset, DIVRR retains $4.50\pm0.20$ entries
versus MemoryEQA's $73.53\pm1.30$, a \textbf{94\%} reduction, while its
$5.705\pm0.061$\,s latency remains close to Explore-EQA's $5.510\pm0.069$\,s.

\begin{table}[!t]
  \centering
  \small
  \setlength{\tabcolsep}{5pt}
  \caption{Zero-shot performance on the static HM-EQA dataset.}
  \label{table:hmeqa}
  \begin{tabular}{lcccc}
    \toprule
    \textbf{Method} & \textbf{Acc} & \textbf{Time} & \textbf{Mem} & \textbf{Step} \\
    \midrule
    Explore-EQA & 54.6 & \textbf{4.36} & -- & 0.75 \\
    Fine-EQA & 58.2 & 5.38 & -- & 0.70 \\
    Graph-EQA & \underline{60.4} & 9.14 & \underline{6.04} & \textbf{0.40} \\
    MemoryEQA & 56.6 & 27.22 & 33.26 & \underline{0.60} \\
    DIVRR & \textbf{63.8} & \underline{4.66} & \textbf{2.51} & 0.64 \\
    \bottomrule
  \end{tabular}
\end{table}

\paragraph*{Results on HM-EQA}
DIVRR reaches 63.8\% accuracy (Table~\ref{table:hmeqa}), exceeding Graph-EQA by
3.4 points with \textbf{58\% fewer memory entries}, and MemoryEQA by 7.2 points
with \textbf{92\% fewer entries}. Its 0.64 extra sensing steps per waypoint is
lower than Explore-EQA and Fine-EQA but higher than Graph-EQA and MemoryEQA.

\subsection{Active-perception controls under a common sensing cap}
\label{sec:active_controls}
We compare question-guided view augmentation with random views and waiting on the
dynamic subset under a common cap of at most three extra observations. The four
controls share the same FBE explorer: (i) \textbf{FBE} (no extra sensing);
(ii) \textbf{Random}, which acquires extra views under the same cap in random
directions, isolating the effect of question conditioning; (iii) \textbf{Wait-3},
which pauses and observes a detected person for three additional frames at each
refinement trigger, emulating a ``stop and wait'' strategy; and (iv)
\textbf{FBE+VR} (ours). To isolate the effect of the extra-view policy, room
detection is disabled throughout this comparison, so column~(iv) measures the
\textbf{+AM+VR} configuration of Table~\ref{table:ablation}.
All methods on the dynamic subset pause to observe detected people for
three frames; Wait-3 additionally waits at refinement triggers.

\begin{table}[!t]
  \centering
  \caption{Controlled active-perception comparisons on the dynamic subset
  (three-seed mean$\pm$std), with a common cap of at most three extra observations.}
  \label{tab:active-controls}
  \setlength{\tabcolsep}{3pt}
  \begin{tabular}{lcccc}
    \toprule
    & \textbf{FBE} & \textbf{Random} & \textbf{Wait-3} & \textbf{FBE+VR} \\
    & & & & \textbf{(no RD)} \\
    \midrule
    Acc.\ (\%) $\uparrow$ & $47.8{\pm}1.05$ & $49.1{\pm}1.02$ & $48.3{\pm}1.23$ & $\mathbf{50.9{\pm}0.93}$ \\
    Extra obs.\ $\downarrow$ & 0 & $\le 3$ & $\le 3$ & $\le 3$ \\
    Time (s) $\downarrow$ & $\mathbf{5.50{\pm}0.13}$ & $5.69{\pm}0.15$ & $5.66{\pm}0.10$ & $5.72{\pm}0.08$ \\
    \bottomrule
  \end{tabular}
\end{table}

As shown in Table~\ref{tab:active-controls}, Wait-3 reaches 48.3\% accuracy,
compared with 47.8\% for FBE.
FBE+VR is 3.1 points above FBE and 1.8 points above Random, with a 0.22\,s increase
over FBE. Under the common sensing cap, these comparisons favor question-guided
view selection over random extra views and the evaluated three-frame waiting policy.

\begin{table}[!t]
  \centering
  \caption{Ablation of DIVRR modules and exploration backbones on DynHuman-EQA
  (\textit{all / dynamic / static}).}
  \label{table:ablation}
  \setlength{\tabcolsep}{4pt}
  \renewcommand{\arraystretch}{1.1}
  \resizebox{\linewidth}{!}{%
  \begin{tabular}{llcc|cc|cc}
    \toprule
    \textbf{Category} & \textbf{Variant}
    & \multicolumn{2}{c|}{\textbf{all}}
    & \multicolumn{2}{c|}{\textbf{dynamic}}
    & \multicolumn{2}{c}{\textbf{static}} \\
    \cmidrule(lr){3-4}\cmidrule(lr){5-6}\cmidrule(lr){7-8}
    & & \textbf{Acc} & \textbf{Mem} & \textbf{Acc} & \textbf{Mem} & \textbf{Acc} & \textbf{Mem} \\
    \midrule
    \multirow{4}{*}{\parbox[c]{1.2cm}{\centering\textbf{Module}}}
      & baseline & 46.27 & -- & 42.00 & -- & 50.55 & -- \\
      & +AM & 51.82 & \textbf{6.50} & 47.64 & \textbf{2.6} & \underline{56.00} & \textbf{10.4} \\
      & +AM+VR & \underline{52.91} & 10.30 & \underline{50.55} & 5.4 & 55.27 & 15.2 \\
      & +AM+VR+RD & \textbf{56.64} & \underline{9.45} & \textbf{55.09} & \underline{4.5} & \textbf{58.18} & \underline{14.4} \\
    \midrule
    \multirow{4}{*}{\parbox[c]{1.2cm}{\centering\textbf{Explor.}}}
      & RE & 33.91 & 9.80 & 33.27 & 5.1 & 34.55 & 14.5 \\
      & GOE & 36.18 & 10.85 & 33.27 & 6.3 & 39.09 & 15.4 \\
      & FBE & \underline{51.82} & \textbf{6.50} & \underline{47.64} & \textbf{2.6} & \underline{56.00} & \textbf{10.4} \\
      & FBE+VR+RD & \textbf{56.64} & \underline{9.45} & \textbf{55.09} & \underline{4.5} & \textbf{58.18} & \underline{14.4} \\
    \bottomrule
  \end{tabular}%
  }
\end{table}

\subsection{Ablation analysis}
\label{sec:ablation}
Table~\ref{table:ablation} ablates DIVRR on DynHuman-EQA, covering module
contributions and exploration backbones.

\paragraph*{Effects of memory admission and exploration modules}
Adding Adaptive Memory (AM) improves accuracy from 46.27\% to 51.82\% with compact
memory. View Refinement (VR) increases overall accuracy to 52.91\%, a gain of
1.09 percentage points. Its effect differs across subsets: dynamic accuracy rises
from 47.64\% to 50.55\%, while static accuracy decreases from 56.00\% to 55.27\%.
Adding Room Detection (RD) achieves
the best accuracy of 56.64\% while reducing memory from 10.30 to 9.45 overall (and
from 5.4 to 4.5 on the dynamic subset) compared with +AM+VR.
RD changes the exploration policy (Sec.~\ref{sec:rd}) while retaining the same
memory admission rule.

\paragraph*{Impact of exploration backbones}
With AM enabled for all exploration backbones, Random Exploration (RE) and
Goal-Oriented Exploration (GOE)~\cite{chaplot2020object} achieve
lower accuracy despite similar entry counts. FBE provides a stronger
foundation with 51.82\% accuracy, while the full FBE-based configuration with VR
and RD reaches 56.64\%. This comparison measures the combined contribution of VR
and RD on top of FBE with AM.

\begin{table}[!t]
  \centering
  \caption{Ablation of local VLM backbones within DIVRR on DynHuman-EQA
  (\textit{all / dynamic / static}).}
  \label{table:vlm_ablation}
  \setlength{\tabcolsep}{5pt}
  \renewcommand{\arraystretch}{1.1}
  \resizebox{\linewidth}{!}{%
  \begin{tabular}{lcc|cc|cc}
    \toprule
    \textbf{VLM backbone}
    & \multicolumn{2}{c|}{\textbf{all}}
    & \multicolumn{2}{c|}{\textbf{dynamic}}
    & \multicolumn{2}{c}{\textbf{static}} \\
    \cmidrule(lr){2-3}\cmidrule(lr){4-5}\cmidrule(lr){6-7}
    & \textbf{Acc} & \textbf{Mem} & \textbf{Acc} & \textbf{Mem} & \textbf{Acc} & \textbf{Mem} \\
    \midrule
    Prismatic-7B            & 49.00             & \underline{10.05} & 46.00             & 5.3             & 52.00             & \underline{14.8} \\
    Janus-Pro-7B            & \underline{51.36} & 10.35             & \underline{46.91} & \underline{5.1} & \underline{55.82} & 15.6             \\
    Qwen2.5-VL-7B (default) & \textbf{56.64}    & \textbf{9.45}    & \textbf{55.09}    & \textbf{4.5}    & \textbf{58.18}    & \textbf{14.4}    \\
    \bottomrule
  \end{tabular}%
  }
\end{table}

\paragraph*{Sensitivity to the VLM backbone}
We evaluate backbone sensitivity by replacing
Qwen2.5-VL-7B inside DIVRR with Prismatic-7B and Janus-Pro-7B, keeping every other
component fixed. As Table~\ref{table:vlm_ablation} shows, the admitted memory counts remain
close across the three backbones (9.45--10.35 overall, 4.5--5.3 dynamic, and
14.4--15.6 static), while accuracy varies more substantially. Qwen2.5-VL-7B is the
strongest on every split and is used as the default throughout. Because replacing
the backbone affects both relevance scoring and final answer generation, this
experiment measures end-to-end backbone sensitivity across both stages.

\begin{table}[!t]
\centering
\caption{Sensitivity to the admission threshold $\tau_{\mathrm{mem}}$ and the
refinement threshold $\tau_{\mathrm{rot}}$ on the dynamic subset of DynHuman-EQA.
}
\label{tab:threshold_sensitivity}
\setlength{\tabcolsep}{5pt}
\begin{tabular}{ccccc}
\toprule
$\tau_{\mathrm{mem}}$ & $\tau_{\mathrm{rot}}$ & Acc(\%)$\uparrow$ & Mem$\downarrow$ & Time$\downarrow$ \\
\midrule
0.7 & 0.5 & 53.82 & 7.6 & 6.41 \\
0.7 & 0.6 & 53.27 & 6.5 & 5.88 \\
0.8 & 0.5 & 54.55 & 5.6 & 6.45 \\
0.8 & 0.6 & \textbf{55.09} & 4.5 & 5.70 \\
0.8 & 0.7 & 53.45 & 3.6 & \textbf{5.34} \\
0.9 & 0.5 & 52.36 & 3.4 & 6.82 \\
0.9 & 0.6 & 52.73 & 2.8 & 6.20 \\
0.9 & 0.7 & 51.45 & \textbf{2.3} & 5.72 \\
\bottomrule
\end{tabular}
\end{table}

\paragraph*{Threshold sensitivity}
Table~\ref{tab:threshold_sensitivity} sweeps $\tau_{\mathrm{mem}}$ and
$\tau_{\mathrm{rot}}$ on the dynamic subset. The two thresholds play distinct and
interpretable roles. Lowering $\tau_{\mathrm{mem}}$ admits more observations and
enlarges memory, whereas raising it too aggressively discards useful evidence; a
lower $\tau_{\mathrm{rot}}$ triggers refinement more often and increases latency,
while a higher one reduces sensing cost at the risk of leaving ambiguous
observations unrefined. The default setting $\tau_{\mathrm{mem}}{=}0.8$,
$\tau_{\mathrm{rot}}{=}0.6$ gives the best accuracy while keeping memory compact
and latency low. Across the tested range, \emph{every} configuration scores
between 51.45\% and 55.09\%, and therefore stays above
the strongest dynamic baseline mean of 45.00\%, demonstrating robustness to the
threshold choice within this range.

\subsection{Reliability of the relevance score}
\label{sec:reliability}
To test whether $s_t$ is a reliable selection signal, we run a pose-grounded analysis
against the reference evidence poses recorded during question generation (definitions
in the supplementary material). Grouping observations by DIVRR's decision bands, the
nearest-reference pose alignment rate rises monotonically with $s_t$:
8.3\% below $\tau_{\mathrm{rot}}$, 31.6\% inside the refinement band, and 68.9\% above
$\tau_{\mathrm{mem}}$. This independent geometric check indicates that high scores
align with question-specific evidence viewpoints. The refinement band has
intermediate geometric alignment between the two other score ranges.

\section{Conclusion and Limitations}
\label{sec:conclusion}
We presented \textbf{DIVRR}, a training-free active evidence-acquisition framework for
EQA in dynamic human-populated scenes, with \textbf{DynHuman-EQA}, a
dynamic/static benchmark with shared scene layouts. RD and relevance-triggered view
augmentation guide exploration and additional observation, while selective memory
admission reduces stored entries. Experiments on DynHuman-EQA and HM-EQA show
gains in accuracy and memory efficiency.

DIVRR targets human-induced perceptual dynamics caused by moving people, transient
occlusions, and viewpoint-dependent visibility. Long-term object-state changes,
scene-layout changes, and complex social behaviors remain outside its current scope.
Its VLM-derived relevance scores and fixed thresholds depend on calibration and can
fail under severe ambiguity or domain shift, when useful candidate views are
unavailable, or when evidence conflicts over long horizons. The evaluation is
conducted entirely in simulation; physical deployment remains to be evaluated under
robot motion constraints, sensor noise, and safety-aware interaction with people.
Promising directions for future work include adaptive thresholds that
adjust to VLM calibration, explicit temporal-consistency checks across waypoints to
reconcile conflicting evidence, and richer classes of dynamic events beyond human
motion.

\bibliographystyle{IEEEtran}
\bibliography{refs}

\end{document}